\documentclass[letterpaper, 10 pt, conference]{ieeeconf}  

\IEEEoverridecommandlockouts                              

\usepackage{booktabs}
\usepackage{multirow}
\usepackage{graphicx}
\usepackage{comment}
\usepackage[ruled,vlined]{algorithm2e}
\usepackage{wrapfig}
\usepackage{multicol}
\usepackage{amssymb}
\usepackage{mathtools}
\usepackage{amsmath}
\usepackage{array}
\usepackage{arydshln} 
\usepackage[T1]{fontenc}
\usepackage[outline]{contour}
\usepackage{xcolor}
\definecolor{darkdarkblue}{RGB}{25, 50, 100}
\definecolor{darkblue}{RGB}{31, 78, 121}
\definecolor{darkgreen}{RGB}{56, 87, 35}
\definecolor{darkred}{RGB}{131, 38, 41}
\usepackage{subcaption}
\usepackage{placeins}
\usepackage{hyperref}
\hypersetup{
    colorlinks=true,
    linkcolor=black,
    citecolor=darkdarkblue,
    urlcolor=blue,
    bookmarks=false,
    hypertexnames=true
}

\title{\LARGE \bf
Coarse-to-Fine Imitation Learning:\\Robot Manipulation from a Single Demonstration
}

\author{Edward Johns\,$^*$
\thanks{$^*$\,Edward Johns is with the Robot Learning Lab at Imperial College London. This work was supported by the Royal Academy of Engineering under the Research Fellowship scheme.}%
}

\begin{document}

\maketitle
\thispagestyle{empty}
\pagestyle{empty}

\begin{abstract}

We introduce a simple new method for visual imitation learning, which allows a novel robot manipulation task to be learned from a single human demonstration, without requiring any prior knowledge of the object being interacted with. Our method models imitation learning as a state estimation problem, with the state defined as the end-effector's pose at the point where object interaction begins, as observed from the demonstration. By then modelling a manipulation task as a coarse, approach trajectory followed by a fine, interaction trajectory, this state estimator can be trained in a self-supervised manner, by automatically moving the end-effector's camera around the object. At test time, the end-effector moves to the estimated state through a linear path, at which point the original demonstration's end-effector velocities are simply replayed. This enables convenient acquisition of a complex interaction trajectory, without actually needing to explicitly learn a policy. Real-world experiments on 8 everyday tasks show that our method can learn a diverse range of skills from a single human demonstration, whilst also yielding a stable and interpretable controller. Videos are available at:\\ \textbf{{\href{http://www.robot-learning.uk/coarse-to-fine-imitation-learning}{www.robot-learning.uk/coarse-to-fine-imitation-learning}}}.

\end{abstract}


\section{INTRODUCTION}

The goal of imitation learning can be described as teaching a robot how to perform a novel task from human demonstrations, whilst minimising two criteria: (1) the amount of physical interaction required of the human, and (2) the amount of prior task knowledge required by the algorithm. Existing solutions typically fail in one or both of these. Methods that learn from demonstrations alone, such as behavioural cloning \cite{zhang2018deep}, require a large number of demonstrations. Methods that bootstrap from demonstrations with additional self-exploration, such as reinforcement learning \cite{vecerik2017leveraging}, require environment resetting via manual intervention or task-specific apparatus. Methods that transfer knowledge from other similar tasks, such as meta-imitation learning \cite{finn2017one}, require prior knowledge of the task family. And methods that analytically model tasks in a manually-defined state space, such as dynamic movement primitives \cite{pastor2009learning}, require prior knowledge of the specific task. In this paper, we introduce a simple new method which addresses both these criteria: our method learns everyday robot manipulation tasks from just a single human demonstration, without requiring any prior knowledge of the object which is being interacted with.

\begin{figure}
\centering
\includegraphics[width=0.8\linewidth]{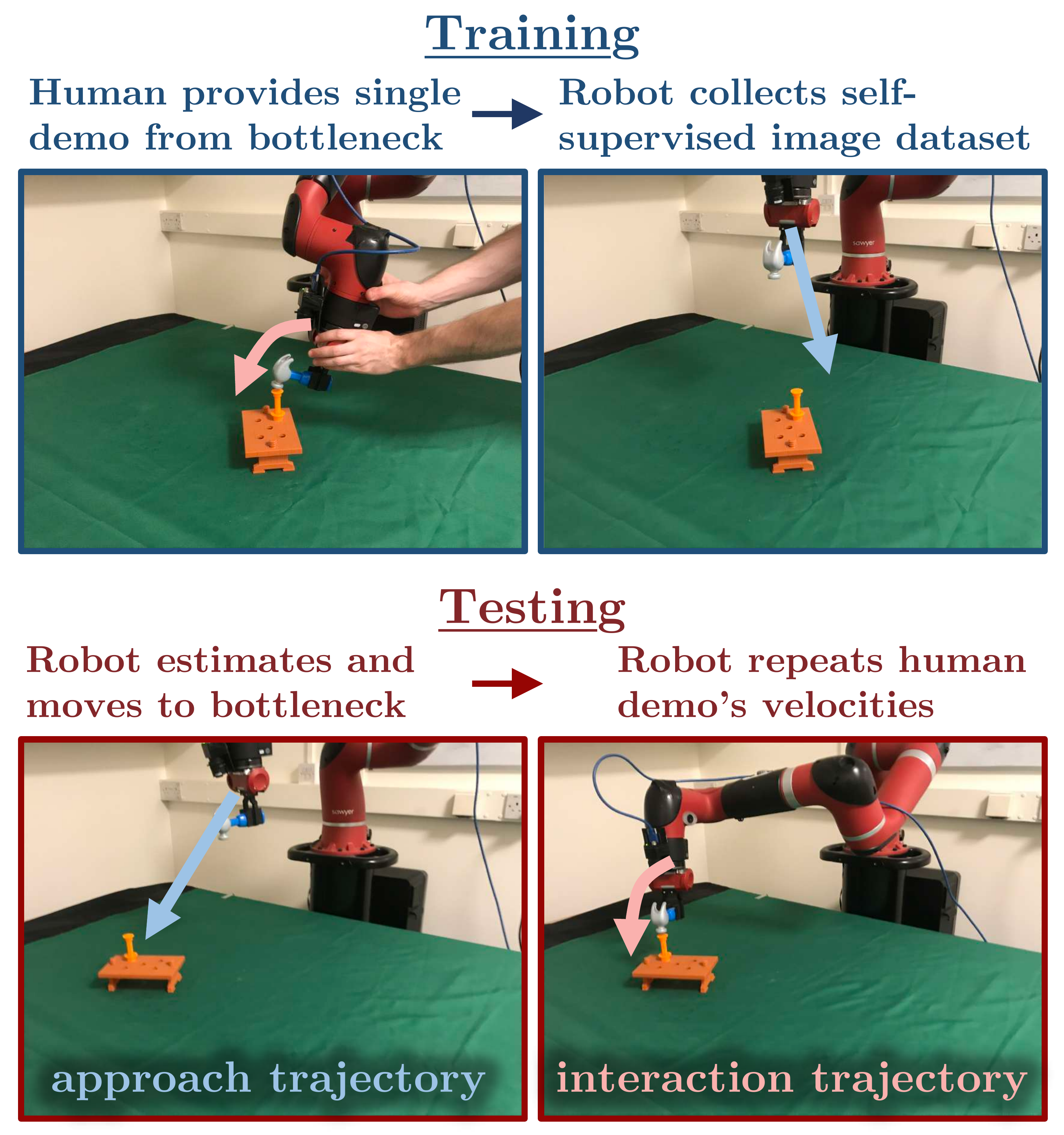}
\caption{We model a robot manipulation task as a coarse \textit{approach trajectory}, followed by a fine \textit{interaction trajectory}. Training for the approach trajectory is self-supervised, and the interaction trajectory needs only a single demonstration. Here, we illustrate teaching the robot how to use a hammer.}
\label{fig:overview}
\end{figure}

Our method \textbf{models the robot's motion as a coarse-to-fine trajectory}, as shown in Figure \ref{fig:overview}. In typical object manipulation tasks, we observe that the end-effector first \textit{approaches} the object's bottleneck in a \textit{coarse} manner through free space, and then physically \textit{interacts} with the object in a \textit{fine} manner. We can thus model the approach analytically with a simple linear path. And therefore, rather than requiring demonstrations in this space, a camera mounted to the end-effector can be moved around the object automatically to collect observations from multiple viewpoints, thus achieving self-supervised generalisation across the task space.

Building on this foundation, our method then \textbf{models imitation learning as a state estimation problem}. Typically, state estimation for robot manipulation requires prior knowledge of the object, such as a pre-defined pose estimator \cite{pastor2009learning} or fiducial markers  \cite{englert2018learning, tsai2021droid}. But we show how to achieve this with a novel object, by using the demonstration itself to define a state in 3D space: the object's \textit{bottleneck}. Intuitively, the bottleneck represents the end-effector's pose at the point where object interaction should begin, as observed from the demonstration, which the robot should then attempt to reach during testing. A bottleneck pose predictor can then be trained using self-supervised learning, as introduced above. During testing, the robot estimates the bottleneck pose and moves the end-effector directly there. This then leads to the \textbf{crux of our method}: if the robot can reach the bottleneck pose with sufficient accuracy, we propose that the robot can then just replay the original demonstration's end-effector velocities from the bottleneck onwards, with the simplest form of behavioural cloning. This enables convenient acquisition of a complex interaction trajectory without needing to explicitly learn a policy. And since the robot explores only in free space, and not during object interaction, we avoid the need for the repeated environment resetting which reinforcement learning suffers from.

In summary, we propose \textbf{Coarse-to-Fine Imitation Learning}, which combines the strengths of both analytical modelling and machine learning, for a data-efficient yet flexible framework. Furthermore, since machine learning is used only for pose estimation and not for policy learning, the controller itself is analytical, so we provide stability and interpretability in contrast to many recent methods in the field which adopt black-box, end-to-end policy learning. In real-world experiments, we study a number of different implementations of our method, and show how it can learn a diverse range of everyday tasks. Videos are available at:\\ \textbf{{\href{http://www.robot-learning.uk/coarse-to-fine-imitation-learning}{www.robot-learning.uk/coarse-to-fine-imitation-learning}}}.


\section{Related Work}

Whilst there is significant prior work on imitation learning using manually-defined low-dimensional state representations \cite{pastor2009learning, englert2018learning, tsai2021droid}, these require task-specific prior knowledge, such as pre-defined pose estimators, or artificial experimental setups, such as fiducial markers. Therefore, in this section, we focus on methods where a robot can learn a new task from a human without requiring state-space engineering, particularly through visual observations.

\textbf{Behavioural cloning methods} use supervised learning to map observations to actions. This is often achieved through end-to-end learning, where a range of experimental setups have been designed to capture actions \cite{zhang2018deep, rahmatizadeh2018vision, florence2019self, young2020visual, james2017transferring, di2020safari}. Image representations can be regularised through keypoint-based architectures \cite{zhang2018deep} or self-supervision \cite{florence2019self}. Similar to our coarse-to-fine formulation, \cite{paradis2020intermittent} combine planning with behavioural cloning. However, since all these methods explicitly learn an end-to-end policy, they require a large number of human demonstrations to achieve generalisation, whereas our method requires only a single demonstration.

\textbf{Exploration-based methods} use self-exploration of the environment to learn a policy, bootstrapping from the original demonstrations. This can be done by pre-training the policy \cite{rajeswaran2018learning}, learning a residual policy \cite{johannink2019residual, silver2018residual}, pre-loading the replay buffer \cite{vecerik2017leveraging}, adding a behavioural cloning loss to the actor \cite{nair2018overcoming}, or restricting the exploration space \cite{thananjeyan2020safety, tsai2020constrained}. A reward function can be inferred from a goal image \cite{schoettler2020deep}, inverse reinforcement learning \cite{finn2016guided}, active learning \cite{singh2019end}, or contrastive learning \cite{sermanet2018time-contrastive}. Generative methods can attempt to reconstruct the demonstrations \cite{ho2016generative, zhu2018reinforcement, schroecker2019generative}. Exploration data can also be used to learn a dynamics model for trajectory optimisation, whose cost is defined by a demonstration \cite{nair2017combining, sieb2020graph}. However, self-exploration methods can be unsafe, and additionally require environment resetting via human intervention or task-specific apparatus, whereas our method does not require any further exploration of object interaction.

\textbf{Transfer learning methods} aim to exploit prior knowledge of similar tasks, to learn a new task from one or a few demonstrations. This can be achieved by supervised learning across a task family \cite{duan2017one}, or by jointly learning a task embedding and policy \cite{james2018task}. Meta-learning an initialisation of network parameters for fast adaptation \cite{finn2017model} can be deployed for behavioural cloning \cite{finn2017one} and learning from observations of humans \cite{yu2018one}. However, the efficient adaptation in these methods comes at the cost of requiring prior training on similar tasks to the new task being learned, whereas our method assumes no prior knowledge of the task family.

Our method is also related to \textbf{visual servoing}. \cite{yu2019siamese, puang2020kovis} attempt to align the robot's current image observation with a goal image, but require task-specific controllers to be manually defined for object interaction, whereas our method can learn from just a demonstration. \cite{argus2020flowcontrol} use visual servoing to directly track a demonstration, but require close initial alignment and only evaluate on tasks with very limited object interaction, whereas our method generalises across a wide task space and can facilitate complex interaction trajectories.


\begin{figure*}[!t]
\centering
\includegraphics[width=0.75\linewidth]{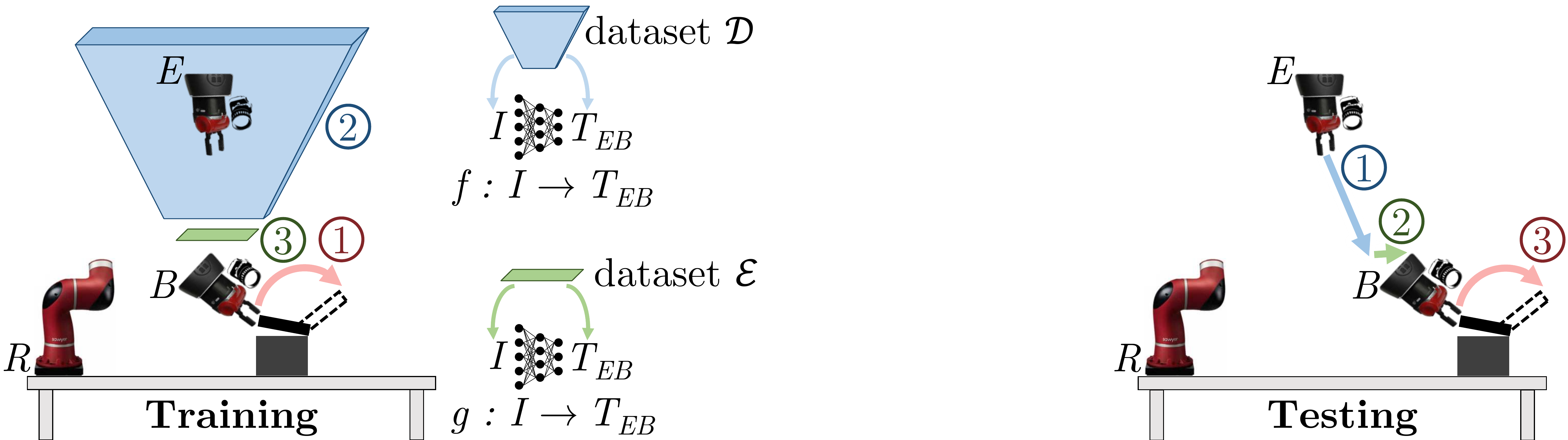}
\caption{Visualising the different stages of training and testing. During training, \textcolor{darkred}{\textcircled{\raisebox{-0.9 pt}{1}}} first a single human demonstration is provided from the bottleneck (e.g. here, opening a lid), \textcolor{darkblue}{\textcircled{\raisebox{-1.2 pt}{2}}} then the robot moves its end-effector and camera around the bottleneck in the blue region, whilst collecting a dataset $\mathcal{D}$ of images $I$ and bottleneck poses $T_{EB}$ to train network $f(I)$, \textcolor{darkgreen}{\textcircled{\raisebox{-0.9 pt}{3}}} then the robot collects a similar dataset $\mathcal{E}$ but in the smaller green region, to train network $g(I)$. During testing, with the object now moved, \textcolor{darkblue}{\textcircled{\raisebox{-0.9 pt}{1}}} first the robot uses $f(I)$ to estimate the bottleneck pose $T_{EB}$ and move towards it, \textcolor{darkgreen}{\textcircled{\raisebox{-0.7 pt}{2}}} then the robot makes a last-inch correction to its estimate using $g(I)$, \textcolor{darkred}{\textcircled{\raisebox{-0.9 pt}{3}}} then the robot replays the demonstration's end-effector velocities.}
\label{fig:frames}
\end{figure*}

\section{Method}
	
\subsection{Coordinate Frames}

We first define the following coordinate frames: the robot's base $R$, the robot's end-effector $E$, and the object's bottleneck $B$, as shown in Figure \ref{fig:frames}. The bottleneck is a ``virtual" frame and does not represent a physical body in the current scene, but represents where the end-effector should be at the point where interaction begins, such that all coarse trajectories converge at the bottleneck. $B$ can considered to be fixed relative to the object. A homogeneous transformation matrix $T_{EB}$ represents frame $B$ expressed in frame $E$.

\subsection{Coarse-to-Fine Trajectories}
\label{sec:coarse-to-fine-trajectories}

We observe that typical manipulation tasks begin with a coarse \textit{approach trajectory}, followed by a fine \textit{interaction trajectory}. During the approach, only the destination is important and not the specific trajectory, whereas during the interaction, the specific trajectory is crucial. We therefore execute the approach trajectory as a simple linear path using inverse kinematics. This does not require human demonstrations, and so we only require a demonstration of the interaction trajectory. From here, two questions now emerge. (1) During testing, what should the robot execute as the interaction trajectory? (2) During testing, where should the interaction trajectory begin? We now introduce two axioms for our assumed environment setup, which answer these.

The \textbf{first axiom} is that, for a controller which outputs actions relative to the end-effector's local frame, those actions should depend only on the relative pose between the end-effector and the object. Therefore, if we record the demonstration's local end-effector velocities from the bottleneck point onwards, then during testing, we can move the end-effector to the bottleneck and then replay those velocities for the interaction trajectory. This answers our first question. Note that since the interaction trajectory is executed open loop, any error in reaching the bottleneck would result in a growing offset from the demonstration, during the interaction trajectory. Therefore, we minimise the length of the interaction trajectory by beginning the demonstration as close to the object as possible, and define the bottleneck pose $T_{RB}$ during training as the demonstration's initial end-effector pose $T_{RE}$. The demonstration should begin no lower than the object's highest point, to avoid collisions during the self-supervised data collection (see Section \ref{sec:self_supervised_learning}).

The \textbf{second axiom} is that the camera's image is a function of the relative pose between the camera and the object. Since we mount the camera rigidly to the end-effector, and the bottleneck is fixed relative to the object, it follows that this image is also a function of the relative pose between the end-effector and the bottleneck, $T_{EB}$. Therefore, we can collect images and associated $T_{EB}$ values by automatically moving the camera around the object, and then train a function which predicts $T_{EB}$ from an image. Then, during testing, the robot can predict the pose of the bottleneck, move the end-effector there, and then execute the interaction trajectory. This answers our second question. Note that in reality, this axiom is an approximation due to observation noise, illumination effects, and variation in background, although these are not significant in our experimental setup and could potentially be mitigated by image augmentation.

\subsection{Self-Supervised Learning}
\label{sec:self_supervised_learning}

\textbf{Data collection}. We define $f(I)$ as the above function to predict $T_{EB}$ from image $I$. After the demonstration, we collect dataset $\mathcal{D}=\{(I, T_{EB})_i\}$, with $T_{EB}$ calculated by:
\begin{equation}
    T_{EB} = T_{ER} \ T_{RB},
    \label{eq:1}
\end{equation}
\noindent where $T_{ER}$ is calculated using forward kinematics, and $T_{RB}$ is the bottleneck pose defined from the demonstration. Note that Equation \ref{eq:1} assumes that the object is in the same pose when $T_{RB}$ is first defined, as when $\mathcal{D}$ is collected. Therefore, if the object moves during the human demonstration, then either it should be re-positioned, or alternatively, $\mathcal{D}$ can be collected before the human demonstration, by first manually setting the bottleneck pose with the end-effector, and later returning to this pose to start the demonstration. Now, since the approach trajectory is a simple linear path and does not require human demonstrations, we can collect $\mathcal{D}$ automatically, by moving the end-effector around in the space above the bottleneck (the blue region in Figure \ref{fig:frames}), to capture the appearance of the object from a range of viewpoints. We do this over a number of trajectories, where each first moves the robot to a random initial pose above the object, and then moves in a straight line at a constant speed towards a pose sampled from near the bottleneck, whilst adding $(I, T_{EB})$ pairs to $\mathcal{D}$. Given dataset $\mathcal{D}$, we then train $f(I)$ as a convolutional neural network using supervised learning. This is done by directly regressing the pose, and therefore we do not require any camera calibration, as long as the camera remains rigidly mounted to the end-effector. 

\textbf{Simplifying the prediction space}. Now, since the specific motion during the approach trajectory is not important, we can constrain the space of bottleneck poses which the trajectory terminates at, and hence simplify the prediction space of the pose estimator $f(I)$. We assume a typical tabletop setup, where the object is limited to horizontal translation and rotation about the vertical axis. To capture only this rotation dimension, we constrain the end-effector poses to be ``vertical" and pointing directly downwards, and allow end-effector rotation only about the vertical axis, both when collecting $\mathcal{D}$, and during the approach trajectory when testing. Given the known height of the bottleneck, this reduces the prediction space of $f(I)$ to three dimensions: two horizontal translations, and one rotation about the vertical. However, all this assumes that the bottleneck is indeed vertical, even though the end-effector's pose at the start of the demonstration may not be. Therefore, we define the bottleneck as having the same position as the demonstration's initial pose, but with a vertical orientation. We then record $R$, the 3D rotation between the bottleneck and this initial pose, which is later applied during testing at the end of the approach trajectory, to re-orientate with the demonstration. Note that during testing, the interaction trajectory can still be 6-DOF despite the bottleneck pose estimation being 3-DOF.

\textbf{Last-inch correction}. Together with collecting dataset $\mathcal{D}$ to train $f(I)$, we also collect a second dataset $\mathcal{E}$ to train a second network $g(I)$, to enable a ``last-inch" correction to the bottleneck pose estimate. $\mathcal{E}$ is collected in a similar manner to $\mathcal{D}$, except that the end-effector's height is fixed at the bottleneck's height throughout data collection, such that each trajectory only moves horizontally with rotation about the vertical (the green region in Figure \ref{fig:frames}). This ensures that $g(I)$ specialises in predictions where the end-effector is already very close to the bottleneck, without images from the remainder of the task space consuming network capacity. During testing, at the end of the approach trajectory, $g(I)$ is then used to make one final estimate of the bottleneck pose. Algorithm \ref{alg:training} summarises the overall training procedure.

\begin{algorithm}
\DontPrintSemicolon
\small
Collect one demo, record initial pose and velocities $\mathcal{U}$ \;
Set bottleneck to initial pose with vertical orientation \;
Set $R$ to rotation between bottleneck and initial pose \;
Set $h$ to height of bottleneck \;
Initialise empty dataset $\mathcal{D}$\;
\For{\textup{\textbf{each} trajectory}}
{
	Move robot to random starting pose\;
	Sample random target pose near to bottleneck\;
	\While{\textup{robot is above} $h$}
	{
		Capture image $I$\ and pose $T_{EB}$ \;
		Store $(I, T_{EB})$ in $\mathcal{D}$\;
		Move robot towards target pose along linear path\;
	}
}
Collect dataset $\mathcal{E}$\;
Train $f(I)$ on $\mathcal{D}$\;
Train $g(I)$ on $\mathcal{E}$\;
\caption{Training algorithm.}
\label{alg:training}
\end{algorithm}

\subsection{Sequential State Estimation}
\label{sec:sequential}
During the approach trajectory, we assume that the object is stationary. Therefore, every prediction of the bottleneck pose is a prediction of the same value. So rather than only using the prediction from the image at the current time step, we can also consider fusing together multiple sequential predictions. First, let us denote $\hat{x}_t$ as 3-dimensional vector representing the horizontal position and vertical orientation of the predicted bottleneck pose at time $t$, such that $\hat{x}_t \equiv f(I_t)$. We also assume a Gaussian uncertainty for this vector, represented as a standard deviation $\hat{\sigma}_t$, which will be discussed below. Then, let us denote $\bar{x}_t$ as the estimate returned by our sequential estimation method, which depends on predictions $\hat{x}_0 \dots \hat{x}_t$. This is also assigned a Gaussian uncertainty $\bar{\sigma}_t$. We now propose two simple methods to calculate $\bar{x}_t$: firstly, by averaging the individual predictions as a \textit{Batch}, and secondly, by \textit{Filtering} the individual predictions.

\textbf{Batch}. In batch estimation, $\bar{x}_t$ is re-estimated at each time step using all previous individual predictions, $\hat{x}_0 \dots \hat{x}_t$. Batch estimation then uses inverse-variance weighting \cite{hartung2011statistical} to combine each $\hat{x}$, which is the optimal estimator under Gaussian uncertainty $\sigma$. The estimate is calculated by:
\begin{equation}
    \bar{x}_t = \frac{\sum_{\tau=0}^{\tau=t} 
    \hat{x}_{\tau} / \hat{\sigma}_{\tau}^2}
    {\sum_{\tau=0}^{\tau=t} 1 / \hat{\sigma}_{\tau}^2}.
    \label{eq:batch_1}
\end{equation}
\textbf{Filtering}. In estimation via filtering, $\bar{x}_t$ is re-estimated at each time step using the estimate from the previous time step, $\bar{x}_{t-1}$, and the prediction at the current time step, $\hat{x}_t$, together with the associated uncertainties:
\begin{equation}
    \bar{x}_t = \frac{(\bar{x}_{t-1} / \bar{\sigma}_{t-1}^2) + (\hat{x}_t / \hat{\sigma}_{t}^2)}{(1/{\bar{\sigma}_{t-1}^2}) + (1/{\hat{\sigma}_{t}^2})},
\end{equation}
\begin{equation}
    \bar{\sigma}_{t}^2 = \frac{1}{(1 / \bar{\sigma}_{t-1}^2) + (1 / \hat{\sigma}_{t}^2)},
\end{equation}
\noindent where the initial $\bar{x}_0$ is set to $\hat{x}_0$, and the initial $\bar{\sigma}_0$ is set to the \textit{Prior} uncertainty, defined below. This is a form of Bayes filter \cite{thrun2002probabilistic}, and is derived from a Kalman filter with zero process noise, since we assume that kinematic errors are insignificant relative to the uncertainty in $f(I)$'s predictions.

\textbf{Estimating Uncertainty}. Both the \textit{Batch} and \textit{Filtering} sequential estimation methods require a Gaussian uncertainty $\hat{\sigma}$ associated with each prediction $\hat{x}$. In our experiments, we investigated three approaches to calculating $\hat{\sigma}$. The first, is to use Dropout to model $f(I)$ as an approximate Bayesian neural network \cite{gal2016dropout}, where we retain Dropout during inference, and set $\hat{\sigma}$ to the standard deviation of the predictions. The second, is to train a function which predicts $\hat{\sigma}$ given $\hat{x}$, in an attempt to capture any dependency between state and uncertainty. For example, it was observed that images captured at a greater distance from the object generally resulted in greater prediction uncertainty. To train this, we took the validation subset of $\mathcal{D}$ used for early stopping of $f(I)$, computed predicted poses for all images, and calculated their errors. Then we trained a small neural network to predict the errors from the predicted poses. The third, is to use $f(I)$'s validation error as a constant, prior uncertainty for all predictions. We refer to these methods in later experiments as \textit{Dropout}, \textit{Predicted}, and \textit{Prior}, respectively.

\subsection{Task Execution}

For testing, the robot executes the approach trajectory by moving along a linear path towards the latest estimate of the bottleneck pose. This pose $T_{RB}$ is calculated by:
\begin{equation}
    T_{RB} = T_{RE} \ T_{EB},
\end{equation}
\noindent where $T_{RE}$ is calculated using forward kinematics, and $T_{EB}$ is the prediction from $f(I)$. When the end-effector reaches the bottleneck's height, $g(I)$ is used to calculate a final estimate of $T_{RB}$. The robot then moves to this pose, and executes the interaction trajectory by replaying the demonstration's end-effector velocities. Algorithm \ref{alg:testing} summarises the overall testing procedure.

\begin{algorithm}
\DontPrintSemicolon
\small
Retrieve $\mathcal{U}$, $R$, $h$, $f(I)$, and $g(I)$ from Algorithm \ref{alg:training}\;
\While{\textup{robot is above} $h$}
{
	Capture image $I$\;
	Predict bottleneck using $f(I)$\;
	Update bottleneck with sequential state estimation\;
	Move robot towards bottleneck along linear path\;
}
Predict bottleneck using $g(I)$\;
Move robot to bottleneck\;
Rotate robot by $R$\;
Execute velocities $\mathcal{U}$\;
\caption{Testing algorithm.}
\label{alg:testing}
\end{algorithm}


\section{Experiments}
\label{sec:experiments}

\subsection{Implementation Details}

We used a basic convolutional neural network for both $f(I)$ and $g(I)$, consisting of four convolutional layers with ReLU and max-pool downsampling, followed by four fully-connected layers with ReLU and Dropout. A mean squared-error loss function was used, with coefficients to balance translation and rotation determined using \cite{kendall2015posenet}, which was calculated using a small dataset and then fixed for all experiments. Early stopping was done with a 20\% validation subset. Real-world experiments used a 7-DOF Sawyer robot operating at 30 Hz, with a camera capturing 64-by-64 RGB images. We found that as the camera is very close to the object by the end of the approach trajectory, higher resolution images were not necessary. The task space was a 40-by-40 cm region on a table below the robot. During training, the object was placed at the centre of the task space, and the robot's approach trajectories began at 50 cm above the table, with initial horizontal positions uniformly sampled from the task space, and initial orientations uniformly sampled over a range of 90 degrees. During each test, the object was placed randomly in the task space, and over a range of 90 degrees relative to its pose during training, and the robot's approach trajectories began at 50 cm above the centre of the task space. For the trajectories used to train $f(I)$, we uniformly sampled a target pose over ranges of $5$ cm and $5^{\circ}$ relative to the bottleneck, with ranges of $2$ cm and $20^{\circ}$ for $g(I)$. 

\subsection{Target Reaching}

We first evaluate a range of methods for bottleneck pose estimation using $f(I)$, including the sequential methods described in Section \ref{sec:sequential}. We created a target reaching task to test the approach trajectory, where the end-effector was first manually moved to a random point above an object to define an artificial bottleneck, with the task then being to estimate that pose and move to it through the approach trajectory. We performed experiments with 7 random objects (different to those for the later imitation learning experiments), using 50 trajectories on each to collect $\mathcal{D}$. We tested each object in 5 different poses, each time setting the end-effector by eye as close to the artificial bottleneck as possible, to record the ground-truth bottleneck pose.

\textbf{Methods}. We implemented 11 different methods for estimating the bottleneck pose. \textit{Oracle} uses the ground-truth bottleneck to reveal the unavoidable error in the position controller. \textit{First Image} deploys $f(I)$ only for the first image in the trajectory, and uses that same estimate throughout the trajectory. \textit{Best Image} continually deploys $f(I)$ for all images, and uses the estimate with the lowest uncertainty from all images captured so far (using \textit{Dropout} or \textit{Predicted}). \textit{Visual Servoing} continually deploys $f(I)$ for all images, and uses the estimate only for the current image without performing sequential estimation. \textit{Batch} and \textit{Filtering} are described in Section \ref{sec:sequential}, and perform sequential estimation by fusing predictions from multiple images along the trajectory (using \textit{Dropout}, \textit{Predicted}, or \textit{Prior}). 


\textbf{Results}.
Table \ref{tab:target_reaching} shows the results, which reveal three insights. Firstly, the best methods for fusing multiple predictions typically outperformed basic visual servoing. This demonstrates the effectiveness of classical, explicit state estimation over implicit state estimation via deep learning, and supports our proposal to model imitation learning as a state estimation problem. Secondly, methods which assume a constant uncertainty for all predictions (\textit{Prior}) typically outperformed those which estimate uncertainty dynamically (\textit{Dropout} and \textit{Predicted}). Since the formulations in Section \ref{sec:sequential} assume Gaussian uncertainties, we can conclude that this assumption introduces a harmful bias since estimating Gaussian uncertainties is challenging with deep learning. Thirdly, filtering methods performed at least as well as batch methods despite discarding past data, which could be explained by the bias towards more recent images captured closer to the object, which are typically more discriminative.

\begin{table}
    \footnotesize
    \centering
    \begin{tabular}{lcccccc}
        \toprule
        \multirow{2}{*} & \multicolumn{2}{c}{Mean Error} & \multicolumn{2}{c}{Min Error} & \multicolumn{2}{c}{Max Error}\\
        \cmidrule(lr){2-3} \cmidrule(lr){4-5} \cmidrule(lr){6-7}
        Estimation Method & Pos & Ori & Pos & Ori & Pos & Ori \\
        \midrule
        Oracle & 0.3 & 0.0 & 0.1 & 0.0 & 0.7 & 0.1 \\
        First Image & 13.9 & 9.0 & 3.7 & \textbf{1.3} & 25.7 & 17.7 \\
        Best Image (Dropout) & \textbf{\underline{4.9}} & 6.4 & \textbf{2.3} & 2.5 & \underline{\textbf{8.2}} & \textbf{10.7} \\
        Best Image (Predicted) & 13.6 & 12.0 & 5.5 & 4.6 & 24.9 & 24.2 \\
        Visual Servoing & 5.7 & 6.6 & 3.4 & 3.1 & \textbf{8.9} & 11.3 \\
        Batch (Prior) & \textbf{5.2} & \textbf{5.8} & \textbf{\underline{1.9}} & 2.0 & 9.8 & \underline{\textbf{10.4}} \\
        Batch (Dropout) & 7.5 & \textbf{5.9} & 2.9 & \underline{\textbf{1.2}} & 12.7 & 12.1 \\
        Batch (Predicted) & 5.8	& 6.4 & 2.3 & 2.2 & 10.2 & 11.3 \\
        Filtering (Prior) & \textbf{5.2} & \underline{\textbf{5.7}} & \textbf{1.9} & 1.8 & \textbf{9.5} & \textbf{10.6}\\
        Filtering (Dropout) & 7.6 & 6.2 & 3.1 & \textbf{1.7} & 12.5 & 12.7 \\
        Filtering (Predicted) & 5.8 & 6.9 & 2.5 & 2.3 & 11.0 & 12.8 \\
        \bottomrule
    \end{tabular}
    \caption{Target reaching results for the approach trajectory, for different bottleneck pose estimation methods. Mean, min and max errors were calculated across 5 poses per object, then averaged across all 7 objects. Positions are in mm, orientations in degrees. The three best results per column (excluding \textit{Oracle}) are in bold, with the best also underlined.}
    \label{tab:target_reaching}
\end{table}

\begin{figure*}[!t]
\centering
  \includegraphics[width=\linewidth]{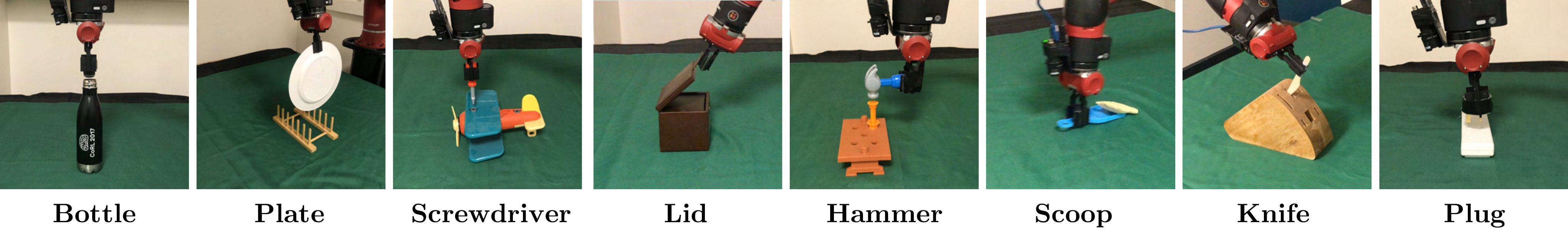}
\caption{The set of 8 everyday tasks we used for real-world imitation learning experiments.}
\label{fig:tasks}
\end{figure*}

\begin{table*}[ht]
    \centering
    \begin{tabular}{lcccccccccc}
        \toprule
        Method & Bottle & Plate & Screwdriver & Lid & Hammer & Scoop & Knife & Plug & Average \\
        \midrule
        Visual Servoing & 65 & 25 & 20 & 65 & 35 & 55 & \textbf{15} & 5 & 35.6 \\
        Visual Servoing + Correction & \textbf{100} & \textbf{95} & \textbf{70} & \textbf{100} & 40 & 95 & 10 & 10 & 65.0 \\
        Filtering & 85 & 25 & 10 & 85 & 50 & 90 & 0 & 10 & 44.4 \\
        Filtering + Correction & \textbf{100} & 80 & 60 & \textbf{100} & \textbf{65} & \textbf{100} & 10 & \textbf{45} & \textbf{70.0} \\
        \bottomrule
    \end{tabular}
    \caption{Real-world imitation learning results, showing \% success rate over 20 object poses for each task. The best result per column is in bold.}
    \label{tab:imitation_learning_results}
\end{table*}

\subsection{Imitation Learning}
The target reaching experiments evaluated the approach trajectory, and we now evaluate our full framework for imitation learning, by introducing human demonstrations and the interaction trajectory. We used 50 trajectories for collecting each of $\mathcal{D}$ and $\mathcal{E}$. Videos are available at:\\ \textbf{{\href{http://www.robot-learning.uk/coarse-to-fine-imitation-learning}{www.robot-learning.uk/coarse-to-fine-imitation-learning}}}.

\textbf{Tasks}. We chose 8 everyday tasks which reflect a range of different challenges, as shown in Figure \ref{fig:tasks}. \textit{Bottle} requires a lid to be placed onto a bottle. \textit{Plate} requires a plate to be inserted into a specific slot in a rack. \textit{Screwdriver} requires a toy screwdriver to be inserted into a screw in a toy aeroplane, and then twisted by 90 degrees. \textit{Lid} requires a lid on a box to be lifted upright. \textit{Knife} requires a wooden knife to be inserted into a knife rack. \textit{Hammer} requires a toy nail to be knocked into a hole using a toy hammer. \textit{Scoop} requires a small bag to be scooped up.  \textit{Plug} requires an electrical plug to be inserted into a socket. For each task, specific success criteria were defined based on the object configuration at the end of the robot's attempt, which were then judged by eye.

\textbf{Methods}. We tested two different methods for the coarse trajectory, each with and without the last-inch correction. First, we evaluated using only the current image to estimate the bottleneck pose at each time step, denoted \textit{Visual Servoing}. Second, we evaluated using \textit{Filtering} with \textit{Prior} uncertainty, since on average this performed at least as well as any other method in the target reaching experiments. We also implemented a method based on residual reinforcement learning following \cite{schoettler2020deep}, using the negative $\ell 1$ distance between the current image and the demonstration's final image as a dense reward, and with the baseline policy trained with behavioural cloning on the single demonstration. However, we found that even with hyper-parameter tuning, this was not able to solve any of the tasks. Whilst this method has previously performed well for insertion tasks using multiple demonstrations \cite{schoettler2020deep}, we found that a single demonstration was not sufficient to constrain exploration, and the $\ell 1$ image distance was often not a meaningful signal for our tasks. We did not compare to behavioural cloning, as this requires more than one demonstration, nor meta-learning, as current methods require prior training on similar tasks.

\textbf{Results}. Table \ref{tab:imitation_learning_results} shows the results for our imitation learning experiments, with 20 object poses tested per task. We see that for the full implementation, using both filtering and the last-inch correction, we are able to learn a range of everyday tasks with encouraging success rates, given that only a single demonstration is required. Three of the tasks were successfully executed in all 20 of the tested object poses. As with the target reaching experiments, sequential state estimation proves to be superior to basic visual servoing. We also see the benefit of the last-inch correction, which significantly increased success rates. Note that since the bottleneck pose estimation is self-supervised, our method can automatically improve in performance with more data collection, without requiring further human demonstrations.

We also notice a large variation in performance across the different tasks. Task difficulty can be defined by two orthogonal properties: first, the required precision to complete the task, and second, the suitability of the object's shape and texture for 2D image-based pose estimation. For example, the \textit{Knife} task was very challenging because not only is the slot in the rack very thin, but the rack itself lacks texture, and appears in an image as just a region of relatively uniform colour, making orientation prediction particularly difficult. The \textit{Plug} task was also challenging, since this is contact-rich and requires significant force to overcome resistance. However, our full method solved this in 9 out of 20 poses of the socket, which is surprisingly high considering only a simple velocity controller is used for contact-rich insertion. However, had the socket not been free to compliantly slide along the table during interaction, the success rates would likely have been lower. Another important observation is how well our method performed on the \textit{Hammer} and \textit{Scoop} tasks. These both require accurate motion at high velocities during object interaction, and are thus well suited to our method of directly replicating the demonstration's velocities. This interaction would be challenging to achieve via explicit policy learning, such as reinforcement learning.

\section{Conclusions}
\label{sec:conclusion}

We have proposed and evaluated \textbf{Coarse-to-Fine Imitation Learning}, a framework which models a robot manipulation task as a coarse, approach trajectory, followed by a fine, interaction trajectory, and models imitation learning as a state estimation problem. We have shown that this can learn a range of novel, real-world, everyday tasks, from a single human demonstration, whereas existing methods typically require either multiple demonstrations, repeated environment resetting, or prior task knowledge. A key component is the ability to simply replay the original demonstration's velocities, enabling convenient acquisition of complex motion, without needing to explicitly learn a policy. Furthermore, the controller is analytical, stable, and interpretable, which typically cannot be said of many visual imitation learning methods today based on end-to-end policy learning. Future work includes improving the bottleneck pose estimator with 3D computer vision, introducing closed-loop control during object interaction, and extending to multi-stage tasks.


\clearpage

\bibliographystyle{IEEEtran}
\bibliography{bibliography}


\end{document}